\documentclass{llncs}
\pdfoutput=1
\usepackage{currfile}
\usepackage[mode=image|tex]{standalone}
\usepackage{xstring} 

\usepackage[T1]{fontenc}
\usepackage[utf8]{inputenc}
\usepackage{csquotes}
\usepackage[main=english]{babel}
\usepackage[hypertexnames=false, unicode, pdfusetitle]{hyperref}

\usepackage{textcomp}
\usepackage{microtype}
\usepackage{etoolbox}
\robustify\bfseries

\usepackage[labelfont=it, font=small, format=plain, labelsep=period]{caption}
\usepackage[skip=3pt]{subcaption}

\usepackage{booktabs}
\usepackage{tabularx}
\usepackage{colortbl}
\usepackage{siunitx}

\usepackage{amsmath}
\usepackage{amssymb}
\usepackage{mathtools}
\usepackage{bm}
\usepackage{xfrac}

\usepackage{url}
\usepackage[capitalise, nameinlink, noabbrev]{cleveref}
\crefformat{equation}{(#2#1#3)}
\usepackage[style=numeric, backend=biber, url=false, maxcitenames=1]{biblatex}

\usepackage{todonotes}

\newcommand{\D}{\mathcal{D}}

\newcommand{\Ell}{\mathcal{L}}

\newcommand{\rv}[1]{\bm{#1}}

\newcommand{\mat}[1]{\bm{#1}}

\DeclareMathOperator{\Oh}{\mathcal{O}}
\DeclareMathOperator{\softmax}{softmax}

\newcommand*{\diff}{\mathop{}\!\mathrm{d}}

\newcommand{\nth}[2][th]{#2^{\text{#1}}}

\DeclareMathOperator{\E}{\mathbb{E}}

\DeclareMathOperator{\p}{p}
\DeclareMathOperator{\q}{q}
\DeclareMathOperator{\KLdiv}{KL}
\DeclareMathOperator{\K}{\mathcal{K}}

\DeclareMathOperator{\Norm}{\mathcal{N}}
\DeclareMathOperator{\Multi}{\mathcal{M}}
\DeclareMathOperator{\Ber}{\mathcal{B}}
\DeclareMathOperator{\Uni}{\mathbb{U}}
\DeclareMathOperator{\Ind}{\mathbb{I}}

\providecommand\given{}
\DeclarePairedDelimiterX{\Cond}[1]{(}{)}{
\renewcommand\given{%
  \nonscript\mkern2mu
  \delimsize\vert
  \nonscript\mkern2mu
  \mathopen{}
  \allowbreak}
#1
}

\makeatletter
\newcommand{\Fun}{\@ifstar\@sfun\@fun}
\newcommand{\@fun}[1]{#1\Cond}
\newcommand{\@sfun}[1]{#1\Cond*}
\makeatother

\newcommand{\Prob}{\p\Cond}

\newcommand{\Variat}{\q\Cond}
\newcommand{\Gaussian}{\Norm\Cond}
\newcommand{\Multinomial}{\Multi\Cond}

\DeclarePairedDelimiterX{\KLdelim}[2]{(}{)}{%
  #1\mkern2mu\delimsize\|\mkern2mu#2%
}
\newcommand{\KL}{\KLdiv\KLdelim}

\DeclarePairedDelimiterXPP{\Moment}[2]{#1}{[}{]}{}{
\renewcommand\given{%
  \nonscript\mkern2mu
  \delimsize\vert
  \nonscript\mkern2mu
  \mathopen{}
  \allowbreak}
#2
}

\DeclarePairedDelimiterX{\Set}[1]{\{}{\}}{

#1
}

\DeclarePairedDelimiterXPP{\pix}[1]{\begingroup\scriptscriptstyle}{(}{)}{\endgroup}{\mkern-2mu#1\mkern-2mu}

\definecolor{tumblue}{HTML}{0065BD}
\definecolor{tumgreen}{HTML}{A2AD00}
\definecolor{tumorange}{HTML}{E37222}
\definecolor{tumivory}{HTML}{DAD7CB}
\definecolor{tumred}{HTML}{E53418} 
\definecolor{tumviolet}{HTML}{69085A} 

\definecolor{tumgray0}{HTML}{000000}
\definecolor{tumgray1}{HTML}{58585A}
\definecolor{tumgray2}{HTML}{9C9D9F}
\definecolor{tumgray3}{HTML}{D9DADB}
\definecolor{tumgray4}{HTML}{FFFFFF}

\definecolor{tumblue0}{HTML}{003359}
\definecolor{tumblue1}{HTML}{005293}
\definecolor{tumblue2}{HTML}{0073CF}
\definecolor{tumblue3}{HTML}{64A0C8}
\definecolor{tumblue4}{HTML}{98C6EA}

\definecolor{tumgreen0}{HTML}{EAF900}
\definecolor{tumgreen1}{HTML}{AEBA00}
\definecolor{tumgreen2}{HTML}{8A9300}
\definecolor{tumgreen3}{HTML}{525800}

\definecolor{tumred0}{HTML}{F23719}
\definecolor{tumred1}{HTML}{CB2E15}
\definecolor{tumred2}{HTML}{90210F}
\definecolor{tumred3}{HTML}{65170B}

\definecolor{tumorange0}{HTML}{F07824}
\definecolor{tumorange1}{HTML}{C9651E}
\definecolor{tumorange2}{HTML}{8E4715}
\definecolor{tumorange3}{HTML}{63320F}

\definecolor{sPetrol}{RGB}{0,153,153}
\definecolor{sStoneDark}{RGB}{60,70,75}
\definecolor{sStoneLight}{RGB}{135,155,170}
\definecolor{sStone}{RGB}{190,205,215}
\definecolor{sSandDark}{RGB}{115,100,90}
\definecolor{sSandLight}{RGB}{170,170,150}
\definecolor{sSand}{RGB}{215,215,205}
\definecolor{sSnow}{RGB}{255,255,255}

\definecolor{sTealDark}{RGB}{0,100,110}
\definecolor{sTealLight}{RGB}{65,170,170}
\definecolor{sBlueDark}{RGB}{0,95,135}
\definecolor{sBlueLight}{RGB}{80,190,215}
\definecolor{sGreenDark}{RGB}{100,125,45}
\definecolor{sGreenLight}{RGB}{170,180,20}
\definecolor{sYellowDark}{RGB}{235,120,10}
\definecolor{sYellowLight}{RGB}{255,185,0}
\definecolor{sRedDark}{RGB}{100,25,70}
\definecolor{sRedLight}{RGB}{175,35,95}

\definecolor{hannah0}{HTML}{d89d1c}
\definecolor{hannah1}{HTML}{d8801c}
\definecolor{hannah2}{HTML}{d8511c}
\definecolor{hannah3}{HTML}{d8bf1c}
\definecolor{hannah4}{HTML}{d8511c}

\title{Data Association with Gaussian Processes}
\author{
    Markus Kaiser\inst{1,2},
    Clemens Otte\inst{1},
    Thomas A. Runkler\inst{1,2},
    Carl Henrik Ek\inst{3}
}
\institute{
    Siemens AG
    \and
    Technical University of Munich
    \and
    University of Bristol
}

\begin{document}
\maketitle

\begin{abstract}
    The data association problem is concerned with separating data coming from different generating processes, for example when data comes from different data sources, contain significant noise, or exhibit multimodality.
    We present a fully Bayesian approach to this problem.
    Our model is capable of simultaneously solving the data association problem and the induced supervised learning problem.
    Underpinning our approach is the use of Gaussian process priors to encode the structure of both the data and the data associations.
    We present an efficient learning scheme based on doubly stochastic variational inference and discuss how it can be applied to deep Gaussian process priors.
\end{abstract}

\section{Introduction}
\label{sec:introduction}
Real-world data often include multiple operational regimes of the considered system, for example a wind turbine or gas turbine~\parencite{hein_benchmark_2017}.
As an example, consider a model describing the lift resulting from airflow around the wing profile of an airplane as a function of the attack angle.
At a low angle the lift increases linearly with attack angle until the wing stalls and the characteristic of the airflow fundamentally changes.
Building a truthful model of such data requires learning two separate models and correctly associating the observed data to each of the dynamical regimes.
A similar example would be if our sensors that measure the lift are faulty in a manner such that we either get an accurate reading or a noisy one.
Estimating a model in this scenario is often referred to as a \emph{data association problem}~\parencite{Bar-Shalom:1987, Cox93areview}, where we consider the data to have been generated by a mixture of processes and we are interested in factorising the data into these components.

\begin{figure}[t]
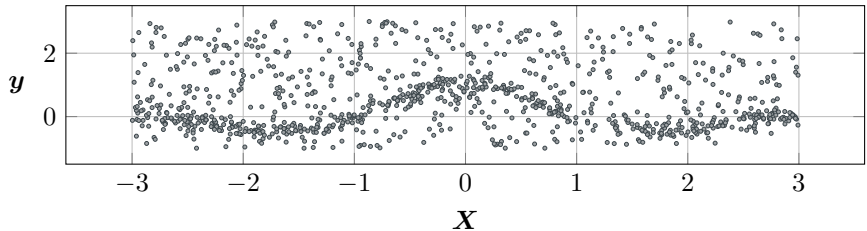

    \centering
    \includestandalone{figures/choicenet_data_intro}
    \caption{
        \label{fig:choicenet_data}
        A data association problem consisting of two generating processes, one of which is a signal we wish to recover and one is an uncorrelated noise process.
    }
\end{figure}

\Cref{fig:choicenet_data} shows an example of faulty sensor data, where sensor readings are disturbed by uncorrelated and asymmetric noise.
Applying standard machine learning approaches to such data can lead to model pollution, where the expressive power of the model is used to explain noise instead of the underlying signal.
Solving the data association problem by factorizing the data into signal and noise gives rise to a principled approach to avoid this behavior.

Early approaches to explaining data using multiple generative processes are based on separating the input space and training local expert models explaining easier subtasks~\parencite{jacobs_adaptive_1991,tresp_mixtures_2001, rasmussen_infinite_2002}.
The assignment of data points to local experts is handled by a gating network, which learns a function from the inputs to assignment probabilities.
However, it is still a central assumption of these models that at every position in the input space exactly one expert should explain the data.
Another approach is presented in~\parencite{bishop_mixture_1994}, where the multimodal regression tasks are interpreted as a density estimation problem.
A high number of candidate distributions is reweighed to match the observed data without modeling the underlying generative process.

In contrast, we are interested in a generative process, where data at the same location in the input space could have been generated by a number of global independent processes.
Inherently, the data association problem is ill-posed and requires assumptions on both the underlying functions and the association of the observations.
In~\parencite{lazaro-gredilla_overlapping_2012} the authors place Gaussian process (GP) priors on the different generative processes which are assumed to be relevant globally.
The associations are modelled via a latent association matrix and inference is carried out using an expectation maximization algorithm.
This approach takes both the inputs and the outputs of the training data into account to solve the association problem.
A drawback is that the model cannot give a posterior estimate about the relevance of the different generating processes at different locations in the input space.
This means that the model can be used for data exploration but additional information is needed in order to perform predictive tasks.
Another approach in~\parencite{bodin_latent_2017} expands this model by allowing interdependencies between the different generative processes and formulating the association problem as an inference problem on a latent space and a corresponding covariance function.
However, in this approach the number of components is a free parameter and is prone to overfitting, as the model has no means of turning off components.

In this paper, we formulate a Bayesian model for the data association problem.
Underpinning our approach is the use of GP priors which encode structure both on the functions and the associations themselves, allowing us to incorporate the available prior knowledge about the proper factorization into the learning problem.
The use of GP priors allows us to achieve principled regularization without reducing the solution space leading to a well-regularized learning problem.
Importantly, we simultaneously solve the association problem for the training data taking both inputs and outputs into account while also obtaining posterior belief about the relevance of the different generating processes in the input space.
Our model can describe non-stationary processes in the sense that a different number of processes can be activated in different locations in the input space.
We describe this non-stationary structure using additional GP priors which allows us to make full use of problem specific knowledge.
This leads to a flexible yet interpretable model with a principled treatment of uncertainty.

The paper has the following contributions:
In \cref{sec:model}, we propose the data association with Gaussian processes model (DAGP).
In \cref{sec:variational_approximation}, we present an efficient learning scheme via a variational approximation which allows us to simultaneously train all parts of our model via stochastic optimization and show how the same learning scheme can be applied to deep GP priors.
We demonstrate our model on a noise separation problem, an artificial multimodal data set, and a multi-regime regression problem based on the cart-pole benchmark in \cref{sec:experiments}.

\section{Data Association with Gaussian Processes}
\label{sec:model}
\begin{figure}[t]
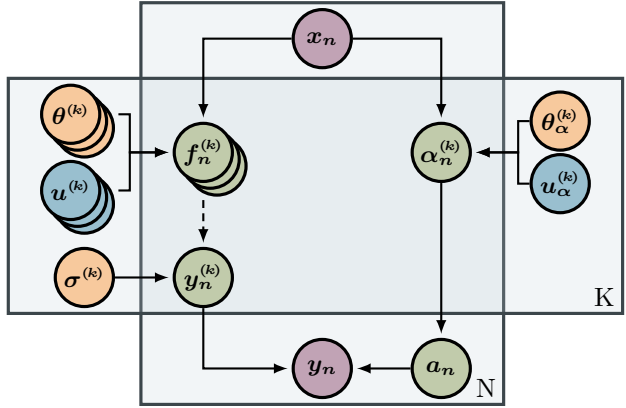

    \centering
    \includestandalone{figures/dynamic_graphical_model}
    \caption{
        \label{fig:dynamic_graphical_model}
        The graphical model of DAGP.
        The violet observations $(\mat{x_n}, \mat{y_n})$ are generated by the latent process (green).
        Exactly one of the $K$ latent functions $f^{\pix{k}}$ and likelihood $\mat{y_n^{\pix{k}}}$ are evaluated to generate $\mat{y_n}$.
        We can place shallow or deep GP priors on these latent function values $\mat{f_n^{\pix{k}}}$.
        The assignment $\mat{a_n}$ to a latent function is driven by input-dependent weights $\mat{\alpha_n^{\pix{k}}}$ which encode the relevance of the different functions at $\mat{x_n}$.
        The different parts of the model are determined by the hyperparameters $\mat{\theta}, \mat{\sigma}$ (yellow) and variational parameters $\mat{u}$ (blue).
    }
\end{figure}
The data association with Gaussian processes (DAGP) model assumes that there exist $K$ independent functions $\Set*{f^{\pix{k}}}_{k=1}^K$, which generate pairs of observations $\D = \Set*{(\mat{x_n}, \mat{y_n})}_{n=1}^N$.
Each data point is generated by evaluating one of the $K$ latent functions and adding Gaussian noise from a corresponding likelihood.
The assignment of the $\nth{n}$ data point to one of the functions is specified by the indicator vector $\mat{a_n} \in \Set*{0, 1}^K$, which has exactly one non-zero entry.
Our goal is to formulate simultaneous Bayesian inference on the functions $f^{\pix{k}}$ and the assignments $\mat{a_n}$.

For notational conciseness, we collect all $N$ inputs as $\mat{X} = \left(\mat{x_1}, \ldots, \mat{x_N}\right)$ and all outputs as $\mat{Y} = \left(\mat{y_1}, \ldots, \mat{y_N}\right)$.
We further denote the $\nth{k}$ latent function value associated with the $\nth{n}$ data point as $\rv{f_n^{\pix{k}}} = \Fun{f^{\pix{k}}}{\mat{x_n}}$ and collect them as $\mat{F^{\pix{k}}} = \left( \rv{f_1^{\pix{k}}}, \ldots, \rv{f_N^{\pix{k}}} \right)$ and $\mat{F} = \left( \mat{F^{\pix{1}}}, \ldots, \mat{F^{\pix{K}}} \right)$.
We refer to the $\nth{k}$ entry in $\mat{a_n}$ as $a_n^{\pix{k}}$ and denote $\mat{A} = \left(\mat{a_1}, \ldots, \mat{a_N}\right)$.

Given this notation, the marginal likelihood of DAGP can be separated into the likelihood, the latent function processes, and the assignment process and is given by,
\begin{align}
\begin{split}
    \label{eq:true_marginal_likelihood}
    \Prob*{\mat{Y} \given \mat{X}} &=
    \int
    \Prob*{\mat{Y} \given \mat{F}, \mat{A}}
    \Prob*{\mat{F} \given \mat{X}}
    \Prob*{\mat{A} \given \mat{X}}
    \diff \mat{A} \diff \mat{F} \\
    \Prob*{\mat{Y} \given \mat{F}, \mat{A}} &=
    \prod_{n=1}^N\prod_{k=1}^K
    \Gaussian*{\mat{y_n} \given \mat{f_n^{\pix{k}}}, \left(\sigma^{\pix{k}}\right)^2}_{^{\displaystyle,}}^{\Fun{\Ind}{a_n^{\pix{k}} = 1}}
\end{split}
\end{align}
where $\sigma^{\pix{k}}$ is the noise level of the $\nth{k}$ Gaussian likelihood and $\Ind$ is the indicator function.

Since we assume the $K$ processes to be independent given the data and assignments, we place independent GP priors on the latent functions
$\Prob*{\mat{F} \given \mat{X}} = \prod_{k=1}^K \Gaussian*{\mat{F^{\pix{k}}} \given \Fun*{\mu^{\pix{k}}}{\mat{X}}, \Fun*{\K^{\pix{k}}}{\mat{X}, \mat{X}}}$.
Our prior on the assignment process is composite.
First, we assume that the $\mat{a_n}$ are drawn independently from multinomial distributions with logit parameters $\mat{\alpha_n} = \left( \alpha_n^{\pix{1}}, \ldots, \alpha_n^{\pix{K}} \right)$.
One approach to specify $\mat{\alpha_n}$ is to assume them to be known a priori and to be equal for all data points~\parencite{lazaro-gredilla_overlapping_2012}.
Instead, we want to infer them from the data.
Specifically, we assume that there is a relationship between the location in the input space $\mathbf{x}$ and the associations.
By placing independent GP priors on $\mat{\alpha^{\pix{k}}}$, we can encode our prior knowledge of the associations by the choice of covariance function
$\Prob*{\mat{\alpha} \given \mat{X}} = \prod_{k=1}^K \Gaussian*{\rv{\alpha^{\pix{k}}} \given \mat{0}, \Fun{\K_\alpha^{\pix{k}}}{\mat{X}, \mat{X}}}$.
The prior on the assignments $\mat{A}$ is given by marginalizing the $\mat{\alpha^{\pix{k}}}$, which, when normalized, parametrize a batch of multinomial distributions,
\begin{align}
\begin{split}
    \label{eq:multinomial_likelihood}
    \Prob*{\mat{A} \given \mat{X}} &=
    \int
    \Multinomial*{\mat{A} \given \Fun{\softmax}{\mat{\alpha}}} \Prob*{\mat{\alpha} \given \mat{X}}
    \diff \rv{\alpha}.
\end{split}
\end{align}
Modelling the relationship between the input and the associations allows us to efficiently model data, which, for example, is unimodal in some parts of the input space and bimodal in others.
A simple smoothness prior will encode a belief for how quickly the components switch across the input domain.

Since the GPs of the $\mat{\alpha^{\pix{k}}}$ use a zero mean function, our prior assumption is a uniform distribution of the different generative processes everywhere in the input space.
If inference on the $\mat{a_n}$ reveals that, say, all data points at similar positions in the input space can be explained by the same $\nth{k}$ process, the belief about $\mat{\alpha}$ can be adjusted to make a non-uniform distribution favorable at this position, thereby increasing the likelihood via $\Prob*{\mat{A} \given \mat{X}}$.
This mechanism introduces an incentive for the model to use as few functions as possible to explain the data and importantly allows us to predict a relative importance of these functions when calculating the posterior of the new observations $\mat{x_\ast}$.

\Cref{fig:dynamic_graphical_model} shows the resulting graphical model, which divides the generative process for every data point in the application of the latent functions on the left side and the assignment process on the right side.
The interdependencies between the data points are introduced through the GP priors on $\rv{f_n^{\pix{k}}}$ and $\rv{\alpha_n^{\pix{k}}}$ and depend on the hyperparameters $\mat{\theta} = \Set*{\mat{\theta^{\pix{k}}}, \mat{\theta_\alpha^{\pix{k}}}, \sigma^{\pix{k}}}_{k=1}^K$.

The priors for the $f^{\pix{k}}$ can be chosen independently to encode different prior assumptions about the underlying processes.
In \cref{subsec:choicenet}, we use different kernels to separate a non-linear signal from a noise process.
Going further, we can also use deep GP as priors for the $f^{\pix{k}}$~\parencite{damianou_deep_2013, salimbeni_doubly_2017}.
Since many real word systems are inherently hierarchical, prior knowledge can often be formulated more easily using composite functions~\parencite{kaiser_bayesian_2018}.

\section{Variational Approximation}
\label{sec:variational_approximation}
Exact inference is intractable in this model.
Instead, we formulate a variational approximation following ideas from~\parencite{hensman_gaussian_2013, salimbeni_doubly_2017}.
Because of the rich structure in our model, finding a variational lower bound which is both faithful and can be evaluated analytically is hard.
To proceed, we formulate an approximation which factorizes along both the $K$ processes and $N$ data points.
This bound can be sampled efficiently and allows us to optimize both the models for the different processes $\Set*{f^{\pix{k}}}_{k=1}^K$ and our belief about the data assignments $\Set*{\mat{a_n}}_{n=1}^N$ simultaneously using stochastic optimization.

\subsection{Variational Lower Bound}
\label{subsec:lower_bound}
As first introduced by~\textcite{titsias_variational_2009}, we augment all GP in our model using sets of $M$ inducing points $\mat{Z^{\pix{k}}} = \left(\mat{z_1^{\pix{k}}}, \ldots, \mat{z_M^{\pix{k}}}\right)$ and their corresponding function values $\mat{u^{\pix{k}}} = \Fun*{f^{\pix{k}}}{\mat{Z^{\pix{k}}}}$, the inducing variables.
We collect them as $\mat{Z} = \Set*{\mat{Z^{\pix{k}}}, \mat{Z_\alpha^{\pix{k}}}}_{k=1}^K$ and $\mat{U} = \Set*{\mat{u^{\pix{k}}}, \mat{u_\alpha^{\pix{k}}}}_{k=1}^K$.
Taking the function $f^{\pix{k}}$ and its corresponding GP as an example, the inducing variables $\mat{u^{\pix{k}}}$ are jointly Gaussian with the latent function values $\mat{F^{\pix{k}}}$ of the observed data by the definition of GPs.
We follow~\parencite{hensman_gaussian_2013} and choose the variational approximation $\Variat*{\mat{F^{\pix{k}}}, \mat{u^{\pix{k}}}} = \Prob*{\mat{F^{\pix{k}}} \given \mat{u^{\pix{k}}}, \mat{X}, \mat{Z^{\pix{k}}}}\Variat*{\mat{u^{\pix{k}}}}$ with $\Variat*{\mat{u^{\pix{k}}}} = \Gaussian*{\mat{u^{\pix{k}}} \given \mat{m^{\pix{k}}}, \mat{S^{\pix{k}}}}$.
This formulation introduces the set $\Set*{\mat{Z^{\pix{k}}}, \mat{m^{\pix{k}}}, \mat{S^{\pix{k}}}}$ of variational parameters indicated in~\cref{fig:dynamic_graphical_model}.
To simplify notation we drop the dependency on $\mat{Z}$ in the following.

A central assumption of this approximation is that given enough well-placed inducing variables $\mat{u^{\pix{k}}}$, they are a sufficient statistic for the latent function values $\mat{F^{\pix{k}}}$.
This implies conditional independence of the $\mat{f_n^{\pix{k}}}$ given $\mat{u^{\pix{k}}}$ and $\mat{X}$.
The variational posterior of a single GP can then be written as,
\begin{align}
\begin{split}
    \Variat*{\mat{F^{\pix{k}}} \given \mat{X}}
    &=
    \int \Variat*{\mat{u^{\pix{k}}}}
    \Prob*{\mat{F^{\pix{k}}} \given \mat{u^{\pix{k}}}, \mat{X}}
    \diff \mat{u^{\pix{k}}}
    \\
    &=
    \int \Variat*{\mat{u^{\pix{k}}}}
    \prod_{n=1}^N \Prob*{\mat{f_n^{\pix{k}}} \given \mat{u^{\pix{k}}}, \mat{x_n}}
    \diff \mat{u^{\pix{k}}},
\end{split}
\end{align}
which can be evaluated analytically, since it is a convolution of Gaussians.
This formulation simplifies inference within single GPs.
Next, we discuss how to handle the correlations between the different functions and the assignment processes.

Given a set of assignments $\mat{A}$, this factorization along the data points is preserved in our model due to the assumed independence of the different functions in~\cref{eq:true_marginal_likelihood}.
The independence is lost if the assignments are unknown.
In this case, both the (a priori independent) assignment processes and the functions influence each other through data with unclear assignments.
Following the ideas of doubly stochastic variational inference (DSVI) presented by~\textcite{salimbeni_doubly_2017} in the context of deep GPs, we maintain these correlations between different parts of the model while assuming factorization of the variational distribution.
That is, our variational posterior takes the factorized form,
\begin{align}
\begin{split}
    \label{eq:variational_distribution}
    \Variat*{\mat{F}, \mat{\alpha}, \mat{U}}
    &= \Variat*{\mat{\alpha}, \Set*{\mat{F^{\pix{k}}}, \mat{u^{\pix{k}}}, \mat{u_\alpha^{\pix{k}}}}_{k=1}^K} \\
    \MoveEqLeft = \prod_{k=1}^K\prod_{n=1}^N \Prob*{\mat{\alpha_n^{\pix{k}}} \given \mat{u_\alpha^{\pix{k}}}, \mat{x_n}}\Variat*{\mat{u_\alpha^{\pix{k}}}}
    \prod_{k=1}^K \prod_{n=1}^N \Prob*{\mat{f_n^{\pix{k}}} \given \mat{u^{\pix{k}}}, \mat{x_n}}\Variat*{\mat{u^{\pix{k}}}}.
\end{split}
\end{align}

Our goal is to recover a posterior for both the generating functions and the assignment of data.
To achieve this, instead of marginalizing $\mat{A}$, we consider the variational joint of $\mat{Y}$ and $\mat{A}$,
\begin{align}
\begin{split}
    \Variat*{\mat{Y}, \mat{A}} &=
    \int
    \Prob*{\mat{Y} \given \mat{F}, \mat{A}}
    \Prob*{\mat{A} \given \mat{\alpha}}
    \Variat*{\mat{F}, \mat{\alpha}}
    \diff \mat{F} \diff \mat{\alpha},
\end{split}
\end{align}
which retains both the Gaussian likelihood of $\mat{Y}$ and the multinomial likelihood of $\mat{A}$ in \cref{eq:multinomial_likelihood}.
A lower bound $\Ell_{\text{DAGP}}$ for the log-joint $\log\Prob*{\mat{Y}, \mat{A} \given \mat{X}}$ of DAGP is given by,
\begin{align}
\begin{split}
    \label{eq:variational_bound}
    \Ell_{\text{DAGP}} &= \Moment*{\E_{\Variat*{\mat{F}, \mat{\alpha}, \mat{U}}}}{\log\frac{\Prob*{\mat{Y}, \mat{A}, \mat{F}, \mat{\alpha}, \mat{U} \given \mat{X}}}{\Variat*{\mat{F}, \mat{\alpha}, \mat{U}}}} \\
    &= \sum_{n=1}^N \Moment*{\E_{\Variat*{\mat{f_n}}}}{\log \Prob*{\mat{y_n} \given \mat{f_n}, \mat{a_n}}}
    + \sum_{n=1}^N \Moment*{\E_{\Variat*{\mat{\alpha_n}}}}{\log \Prob*{\mat{a_n} \given \mat{\alpha_n}}} \\
    &\quad - \sum_{k=1}^K \KL{\Variat*{\mat{u^{\pix{k}}}}}{\Prob*{\mat{u^{\pix{k}}} \given \mat{Z^{\pix{k}}}}}
    - \sum_{k=1}^K \KL{\Variat*{\mat{u_\alpha^{\pix{k}}}}}{\Prob*{\mat{u_\alpha^{\pix{k}}} \given \mat{Z_\alpha^{\pix{k}}}}}.
\end{split}
\end{align}
Due to the structure of~\cref{eq:variational_distribution}, the bound factorizes along the data enabling stochastic optimization.
This bound has complexity $\Fun*{\Oh}{NM^2K}$ to evaluate.

\subsection{Optimization of the Lower Bound}
\label{subsec:computation}
An important property of the variational bound for DSVI~\parencite{salimbeni_doubly_2017} is that taking samples for single data points is straightforward and can be implemented efficiently.
Specifically, for some $k$ and $n$, samples $\mat{\hat{f}_n^{\pix{k}}}$ from $\Variat*{\mat{f_n^{\pix{k}}}}$ are independent of all other parts of the model and can be drawn using samples from univariate unit Gaussians using reparametrizations~\parencite{kingma_variational_2015,rezende_stochastic_2014}.

Note that it would not be necessary to sample from the different processes, since $\Variat*{\mat{F^{\pix{k}}}}$ can be computed analytically~\parencite{hensman_gaussian_2013}.
However, we apply the sampling scheme to the optimization of both the assignment processes $\mat{\alpha}$ and the assignments $\mat{A}$ as for $\mat{\alpha}$, the analytical propagation of uncertainties through the $\softmax$ renormalization and multinomial likelihoods is intractable but can easily be evaluated using sampling.

We optimize $\Ell_{\text{DAGP}}$ to simultaneously recover maximum likelihood estimates of the hyperparameters $\mat{\theta}$, the variational parameters $\Set*{\mat{Z}, \mat{m}, \mat{S}}$, and assignments $\mat{A}$.
For every $n$, we represent the belief about $\mat{a_n}$ as a $K$-dimensional discrete distribution $\Variat*{\mat{a_n}}$.
This distribution models the result of drawing a sample from $\Multinomial*{\mat{a_n} \given \Fun{\softmax}{\mat{\alpha_n}}}$ during the generation of the data point $(\mat{x_n}, \mat{y_n})$.

Since we want to optimize $\Ell_{\text{DAGP}}$ using (stochastic) gradient descent, we need to employ a continuous relaxation to gain informative gradients of the bound with respect to the binary (and discrete) vectors $\mat{a_n}$.
One straightforward way to relax the problem is to use the current belief about $\Variat*{\mat{a_n}}$ as parameters for a convex combination of the $\mat{f_n^{\pix{k}}}$, that is, to approximate $\mat{f_n} \approx \sum_{k=1}^K \Variat*{\mat{a_n^{\pix{k}}}}\mat{\hat{f}_n^{\pix{k}}}$.
Using this relaxation is problematic in practice.
Explaining data points as mixtures of the different generating processes violates the modelling assumption that every data point was generated using exactly one function but can substantially simplify the learning problem.
Because of this, special care must be taken during optimization to enforce the sparsity of $\Variat*{\mat{a_n}}$.

To avoid this problem, we propose using a different relaxation based on additional stochasticity.
Instead of directly using $\Variat*{\mat{a_n}}$ to combine the $\mat{f_n^{\pix{k}}}$, we first draw a sample $\mat{\hat{a}_n}$ from a concrete random variable as suggested by~\textcite{maddison_concrete_2016}, parameterized by $\Variat*{\mat{a_n}}$.
Based on a temperature parameter $\lambda$, a concrete random variable enforces sparsity but is also continuous and yields informative gradients using automatic differentiation.
Samples from a concrete random variable are unit vectors and for $\lambda \to 0$ their distribution approaches a discrete distribution.

Our approximate evaluation of the bound in \cref{eq:variational_bound} during optimization has multiple sources of stochasticity, all of which are unbiased.
First, we approximate the expectations using Monte Carlo samples $\mat{\hat{f}_n^{\pix{k}}}$, $\mat{\hat{\alpha}_n^{\pix{k}}}$, and $\mat{\hat{a}_n}$.
And second, the factorization of the bound along the data allows us to use mini-batches for optimization~\parencite{salimbeni_doubly_2017, hensman_gaussian_2013}.

\subsection{Approximate Predictions}
\label{subsec:predictions}
Predictions for a test location $\mat{x_\ast}$ are mixtures of $K$ independent Gaussians, given by,
\begin{align}
\begin{split}
    \label{eq:predictive_posterior}
    \Variat*{\mat{f_\ast} \given \mat{x_\ast}}
    &= \int \sum_{k=1}^K \Variat*{a_\ast^{\pix{k}} \given \mat{x_\ast}} \Variat*{\mat{f_\ast^{\pix{k}}} \given \mat{x_\ast}} \diff \mat{a_\ast^{\pix{k}}}
    \approx \sum_{k=1}^K \hat{a}_\ast^{\pix{k}} \mat{\hat{f}_\ast^{\pix{k}}}.
\end{split}
\end{align}
The predictive posteriors of the $K$ functions $\Variat*{\mat{f_\ast^{\pix{k}}} \given \mat{x_\ast}}$ are given by $K$ independent shallow GPs and can be calculated analytically~\parencite{hensman_gaussian_2013}.
Samples from the predictive density over $\Variat*{\mat{a_\ast} \given \mat{x_\ast}}$ can be obtained by sampling from the GP posteriors $\Variat*{\mat{\alpha_\ast^{\pix{k}}} \given \mat{x_\ast}}$ and renormalizing the resulting vector $\mat{\alpha_\ast}$ using the $\softmax$-function.
The distribution $\Variat*{\mat{a_\ast} \given \mat{x_\ast}}$ reflects the model's belief about how many and which of the $K$ generative processes are relevant at the test location $\mat{x_\ast}$ and their relative probability.

\subsection{Deep Gaussian Processes}
\label{subsec:deep_gp}
For clarity, we have described the variational bound in terms of a shallow GP.
However, as long as their variational bound can be efficiently sampled, any model can be used in place of shallow GPs for the $f^{\pix{k}}$.
Since our approximation is based on DSVI, an extension to deep GPs is straightforward.
Analogously to~\parencite{salimbeni_doubly_2017}, our new prior assumption about the $\nth{k}$ latent function values $\Prob*{\mat{F^{\prime\pix{k}}} \given \mat{X}}$ is given by,
\begin{align}
\begin{split}
    \Prob*{\mat{F^{\prime\pix{k}}} \given \mat{X}} = \prod_{l=1}^L \Prob*{\mat{F_l^{\prime\pix{k}}} \given \mat{u_l^{\prime\pix{k}}} \mat{F_{l-1}^{\prime\pix{k}}}, \mat{Z_l^{\prime\pix{k}}}},
\end{split}
\end{align}
for an $L$-layer deep GP and with $\mat{F_0^{\prime\pix{k}}} \coloneqq \mat{X}$.
Similar to the single-layer case, we introduce sets of inducing points $\mat{Z_l^{\prime\pix{k}}}$ and a variational distribution over their corresponding function values $\Variat*{\mat{u_l^{\prime\pix{k}}}} = \Gaussian*{\mat{u_l^{\prime\pix{k}}} \given \mat{m_l^{\prime\pix{k}}}, \mat{S_l^{\prime\pix{k}}}}$.
We collect the latent multi-layer function values as $\mat{F^\prime} = \Set{\mat{F_l^{\prime\pix{k}}}}_{k=1,l=1}^{K,L}$ and corresponding $\mat{U^\prime}$ and assume an extended variational distribution,
\begin{align}
\begin{split}
    \label{eq:deep_variational_distribution}
    \Variat*{\mat{F^\prime}, \mat{\alpha}, \mat{U^\prime}}
    &= \Variat*{\mat{\alpha}, \Set*{\mat{u_\alpha^{\pix{k}}}}_{k=1}^K, \Set*{\mat{F_l^{\prime\pix{k}}}, \mat{u_l^{\prime\pix{k}}}}_{k=1,l=1}^{K,L}} \\
    \MoveEqLeft[4] = \prod_{k=1}^K\prod_{n=1}^N \Prob*{\mat{\alpha_n^{\pix{k}}} \given \mat{u_\alpha^{\pix{k}}}, \mat{x_n}}\Variat*{\mat{u_\alpha^{\pix{k}}}}
    \prod_{k=1}^K \prod_{l=1}^L \prod_{n=1}^N \Prob*{\mat{f_{n,l}^{\prime\pix{k}}} \given \mat{u_l^{\prime\pix{k}}}, \mat{x_n}}\Variat*{\mat{u_l^{\prime\pix{k}}}},
\end{split}
\end{align}
where we identify $\mat{f_n^{\prime\pix{k}}} = \mat{f_{n,L}^{\prime\pix{k}}}$.
As the $\nth{n}$ marginal of the $\nth{L}$ layer depends only on the $\nth{n}$ marginal of all layers above sampling from them remains straightforward~\parencite{salimbeni_doubly_2017}.
The marginal is given by,
\begin{align}
\begin{split}
    \Variat{\mat{f_{n,L}^{\prime\pix{k}}}} =
    \int
    \Variat{\mat{f_{n,L}^{\prime\pix{k}}} \given \mat{f_{n,L-1}^{\prime\pix{k}}}}
    \prod_{l=1}^{L-1} \Variat{\mat{f_{n,l}^{\prime\pix{k}}} \given \mat{f_{n,l-1}^{\prime\pix{k}}}}
    \diff \mat{f_{n,l}^{\prime\pix{k}}}.
\end{split}
\end{align}

The complete bound is structurally similar to \cref{eq:variational_bound} and given by,
\begin{align}
\begin{split}
    \label{eq:deep_variational_bound}
    \Ell^\prime_{\text{DAGP}}
    &= \sum_{n=1}^N \Moment*{\E_{\Variat*{\mat{f^\prime_n}}}}{\log \Prob*{\mat{y_n} \given \mat{f^\prime_n}, \mat{a_n}}}
    + \sum_{n=1}^N \Moment*{\E_{\Variat*{\mat{\alpha_n}}}}{\log \Prob*{\mat{a_n} \given \mat{\alpha_n}}} \\
    \MoveEqLeft - \sum_{k=1}^K \sum_{l=1}^L \KL{\Variat{\mat{u_l^{\pix{k}}}}}{\Prob{\mat{u_l^{\pix{k}}} \given \mat{Z_l^{\pix{k}}}}}
    - \sum_{k=1}^K \KL{\Variat*{\mat{u_\alpha^{\pix{k}}}}}{\Prob*{\mat{u_\alpha^{\pix{k}}} \given \mat{Z_\alpha^{\pix{k}}}}}.
\end{split}
\end{align}
To calculate the first term, samples have to be propagated through the deep GP structures.
This extended bound thus has complexity $\Fun*{\Oh}{NM^2LK}$ to evaluate in the general case and complexity $\Fun*{\Oh}{NM^2\cdot\Fun{\max}{L, K}}$ if the assignments $\mat{a_n}$ take binary values.

\section{Experiments}
\label{sec:experiments}
\begin{table}[t]
    \centering
    \caption{
        \label{tab:model_capabilities}
        Comparison of qualitative model capabilities.
        A model has a capability if it contains components which enable it to solve the respective task in principle.
    }
    \scriptsize
    \newcolumntype{Y}{>{\centering\arraybackslash}X}%
    \newcommand{\yes}{\checkmark}
    \newcommand{\no}{--}
    \newcommand{\resultrow}[9]{#1 & #4 & #7 & #3 & #9 & #5 & #6 & #8 \\}
    \begin{tabularx}{\linewidth}{lYYYYYYYY}
        \toprule
        \resultrow{}{Bayesian}{Scalable Inference}{Predictive Posterior}{Data Association}{Predictive Associations}{Multimodal Data}{Separate Models}{Interpretable Priors}
        \midrule
        Experiment & & & & & \cref{tab:choicenet} & \cref{tab:cartpole} & \cref{fig:semi_bimodal} \\
        \midrule
        \resultrow{DAGP (Ours)}{\yes}{\yes}{\yes}{\yes}{\yes}{\yes}{\yes}{\yes}
        \addlinespace
        \resultrow{OMGP \parencite{lazaro-gredilla_overlapping_2012}}{\yes}{\no}{\yes}{\yes}{\no}{\yes}{\yes}{\yes}
        \resultrow{RGPR \parencite{rasmussen_infinite_2002}}{\yes}{\no}{\yes}{\no}{\no}{\yes}{\no}{\yes}
        \resultrow{GPR}{\yes}{\yes}{\yes}{\no}{\no}{\no}{\no}{\yes}
        \addlinespace
        \resultrow{BNN+LV \parencite{depeweg_learning_2016}}{\yes}{\yes}{\yes}{\no}{\no}{\yes}{\no}{\no}
        \resultrow{MDN \parencite{bishop_mixture_1994}}{\no}{\yes}{\yes}{\no}{\no}{\yes}{\no}{\no}
        \resultrow{MLP}{\no}{\yes}{\yes}{\no}{\no}{\no}{\no}{\no}
        \bottomrule
    \end{tabularx}
\end{table}
In this section, we investigate the behavior of the DAGP model.
We use an implementation of DAGP in TensorFlow~\parencite{tensorflow2015-whitepaper} based on GPflow~\parencite{matthews_gpflow_2017} and the implementation of DSVI~\parencite{salimbeni_doubly_2017}.
\Cref{tab:model_capabilities} compares qualitative properties of DAGP and related work.
All models can solve standard regression problems and yield unimodal predictive distributions or, in case of multi-layer perceptrons (MLP), a single point estimate.
Both standard Gaussian process regression (GPR) and MLP do not impose structure which enables the models to handle multi-modal data.
Mixture density networks (MDN)~\parencite{bishop_mixture_1994} and the infinite mixtures of Gaussian processes (RGPR)~\parencite{rasmussen_infinite_2002} model yield multi-modal posteriors through mixtures with many components but do not solve an association problem.
Similarly, Bayesian neural networks with added latent variables (BNN+LV)~\parencite{depeweg_learning_2016} represent such a mixture through a continuous latent variable.
Both the overlapping mixtures of Gaussian processes (OMGP)~\parencite{lazaro-gredilla_overlapping_2012} model and DAGP explicitly model the data association problem and yield independent models for the different generating processes.
However, OMGP assumes global relevance of the different modes.
In contrast, DAGP infers a spacial posterior of this relevance.
We evaluate our model on three problems to highlight the following advantages of the explicit structure of DAGP:

\emph{Interpretable priors give structure to ill-posed data association problems.}
In \cref{subsec:choicenet}, we consider a noise separation problem, where a signal of interest is disturbed with uniform noise.
To solve this problem, assumptions about what constitutes a signal are needed.
The hierarchical structure of DAGP allows us to formulate independent and interpretable priors on the noise and signal processes.

\emph{Predictive associations represent knowledge about the relevance of generative processes.}
In \cref{subsec:semi_bimodal}, we investigate the implicit incentive of DAGP to explain data using as few processes as possible.
Additional to a joint posterior explaining the data, DAGP also gives insight into the relative importance of the different processes in different parts of the input space.
DAGP is able to explicitly recover the changing number of modes in a data set.

\emph{Separate models for independent generating processes avoid model pollution.}
In \cref{subsec:cartpole}, we simulate a system with multiple operational regimes via mixed observations of two different cart-pole systems.
DAGP successfully learns an informative joint posterior by solving the underlying association problem.
We show that the DAGP posterior contains two separate models for the two original operational regimes.

\subsection{Noise Separation}
\label{subsec:choicenet}
\begin{figure}[t]
    \centering
    \captionof{table}{
        \label{tab:choicenet}
        Results on the ChoiceNet data set.
        The gray part of the table shows RMSE results for baseline models from~\parencite{choi_choicenet_2018}.
        For our experiments using the same setup, we report RMSE comparable to the previous results together with MLL.
        Both are calculated based on a test set of 1000 equally spaced samples of the noiseless underlying function.
    }%
    \newcolumntype{H}{>{\setbox0=\hbox\bgroup}c<{\egroup}@{}}
    \newcolumntype{Y}{>{\centering\arraybackslash}X}%
    \newcolumntype{Z}{>{\columncolor{sStone!33}\centering\arraybackslash}X}%
    \begin{tabularx}{\linewidth}{rYYYY|ZZZZHZ}
        \toprule
        Outliers & DAGP & OMGP & DAGP & OMGP & CN & MDN & MLP & GPR & LGPR & RGPR \\
        & \scriptsize MLL & \scriptsize MLL & \scriptsize RMSE & \scriptsize RMSE & \scriptsize RMSE & \scriptsize RMSE & \scriptsize RMSE & \scriptsize RMSE & \scriptsize RMSE & \scriptsize RMSE \\
        \midrule
        0\,\% & \textbf{2.86} & 2.09 & 0.008 & \textbf{0.005} & 0.034 & 0.028 & 0.039 & 0.008 & 0.022 & 0.017 \\
        20\,\% & \textbf{2.71} & 1.83 & 0.008 & \textbf{0.005} & 0.022 & 0.087 & 0.413 & 0.280 & 0.206 & 0.013 \\
        40\,\% & \textbf{2.12} & 1.60 & \textbf{0.005} & 0.007 & 0.018 & 0.565 & 0.452 & 0.447 & 0.439 & 1.322 \\
        60\,\% & 0.874 & \textbf{1.23} & 0.031 & \textbf{0.006} & 0.023 & 0.645 & 0.636 & 0.602 & 0.579 & 0.738 \\
        80\,\% & \textbf{0.126} & -1.35 & 0.128 & 0.896 & \textbf{0.084} & 0.778 & 0.829 & 0.779 & 0.777 & 1.523 \\
        \bottomrule
    \end{tabularx}
    \\[\baselineskip]
    \begin{subfigure}{.37\linewidth}
        \centering
        \includestandalone{figures/choicenet_data_40}
    \end{subfigure}%
    \begin{subfigure}{.315\linewidth}
        \centering
        \includestandalone{figures/choicenet_joint_40}
    \end{subfigure}%
    \begin{subfigure}{.315\linewidth}
        \centering
        \includestandalone{figures/choicenet_attrib_40}
    \end{subfigure}%
    \\
    \begin{subfigure}{.37\linewidth}
        \centering
        \includestandalone{figures/choicenet_data}
    \end{subfigure}%
    \begin{subfigure}{.315\linewidth}
        \centering
        \includestandalone{figures/choicenet_joint}
    \end{subfigure}%
    \begin{subfigure}{.315\linewidth}
        \centering
        \includestandalone{figures/choicenet_attrib}
    \end{subfigure}%
    \captionof{figure}{
        \label{fig:choicenet}
        DAGP on the ChoiceNet data set with 40\,\% outliers (upper row) and 60\,\% outliers (lower row).
        We show the raw data (left), joint posterior (center) and assignments (right).
        The bimodal DAGP identifies the signal perfectly up to 40\,\% outliers.
        For 60\,\% outliers, some of the noise is interpreted as signal, but the latent function is still recovered.
    }
\end{figure}
We consider an experiment based on a noise separation problem.
We apply DAGP to a one-dimensional regression problem with uniformly distributed asymmetric outliers in the training data.
We use a task proposed by~\textcite{choi_choicenet_2018} where we sample $x \in [-3, 3]$ uniformly and apply the function $\Fun{f}{x} = (1 - \delta)(\Fun{\cos}{\sfrac{\pi}{2} \cdot x}\Fun{\exp}{-(\sfrac{x}{2})^2} + \gamma) + \delta \cdot \epsilon$, where $\delta \sim \Fun{\Ber}{\lambda}$, $\epsilon \sim \Fun{\Uni}{-1, 3}$ and $\gamma \sim \Gaussian{0, 0.15^2}$.
That is, a fraction $\lambda$ of the training data, the outliers, are replaced by asymmetric uniform noise.
We sample a total of 1000 data points and use $25$ inducing points for every GP in our model.

Every generating process in our model can use a different kernel and therefore encode different prior assumptions.
For this setting, we use two processes, one with a squared exponential kernel and one with a white noise kernel.
This encodes the problem statement that every data point is either part of the signal we wish to recover or uncorrelated noise.
To avoid pathological solutions for high outlier ratios, we add a prior to the likelihood variance of the first process, which encodes our assumption that there actually is a signal in the training data.

The model proposed in~\parencite{choi_choicenet_2018}, called ChoiceNet (CN), is a specific neural network structure and inference algorithm to deal with corrupted data.
In their work, they compare their approach to the MLP, MDN, GPR, and RGPR models.
We add experiments for both DAGP and OMGP.
\Cref{tab:choicenet} shows results for outlier rates varied from 0\,\% to 80\,\%.
Besides the root mean squared error (RMSE) reported in~\parencite{choi_choicenet_2018}, we also report the mean test log likelihood (MLL).

Since we can encode the same prior knowledge about the signal and noise processes in both OMGP and DAGP, the results of the two models are comparable:
For low outlier rates, they correctly identify the outliers and ignore them, resulting in a predictive posterior of the signal equivalent to standard GP regression without outliers.
In the special case of 0\,\% outliers, the models correctly identify that the process modelling the noise is not necessary, thereby simplifying to standard GP regression.
For high outlier rates, stronger prior knowledge about the signal is required to still identify it perfectly.
\Cref{fig:choicenet} shows the DAGP posterior for an outlier rate of 60\,\%.
While the function has still been identified well, some of the noise is also explained using this process, thereby introducing slight errors in the predictions.

\subsection{Multimodal Data}
\label{subsec:semi_bimodal}
\begin{figure}[t]
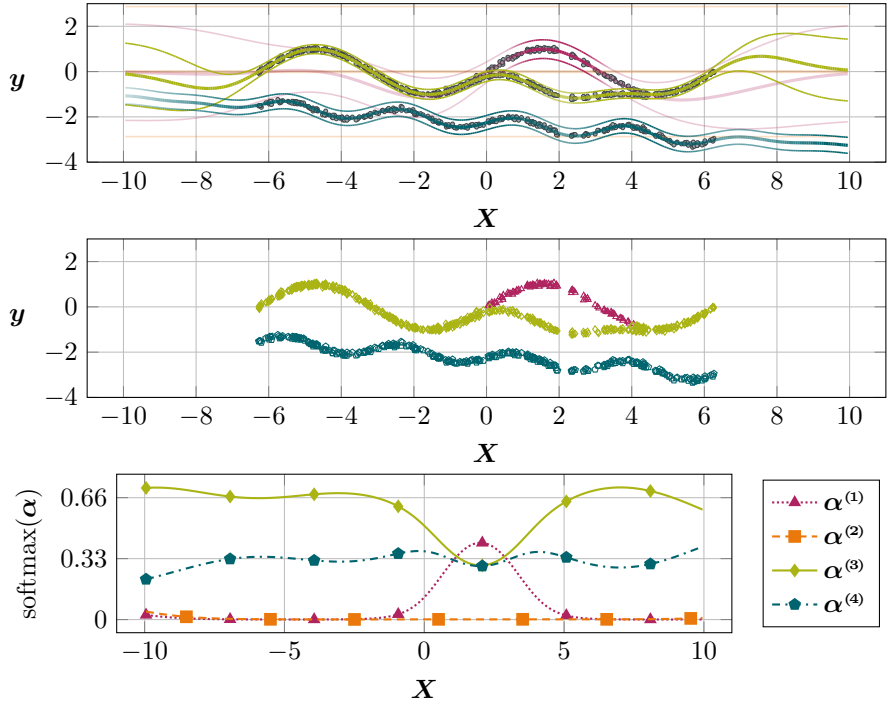

    \centering
    \includestandalone{figures/semi_bimodal_joint}
    \includestandalone{figures/semi_bimodal_attrib}
    \includestandalone{figures/semi_bimodal_attrib_process}
    \caption{
        \label{fig:semi_bimodal}
        The DAGP posterior on an artificial data set with bimodal and trimodal parts.
        The joint predictions (top) are mixtures of four Gaussians weighed by the assignment probabilities $\mat{\alpha}$ (bottom).
        The weights are represented via the opacity of the modes.
        The model has learned that the mode $k = 2$ is irrelevant, that the mode $k = 1$ is only relevant around the interval $[0, 5]$.
        Outside this interval, the mode $k = 3$ is twice as likely as the mode $k = 4$.
        The concrete assignments $\mat{a}$ (middle) of the training data show that the mode $k = 1$ is only used to explain observations where the training data is trimodal.
        The mode $k = 2$ is never used.
    }
\end{figure}
Our second experiment applies DAGP to a multimodal data set.
The data, together with recovered posterior attributions, can be seen in \cref{fig:semi_bimodal}.
We uniformly sample 350 data points in the interval $x \in [-2\pi, 2\pi]$ and obtain $y_1 = \Fun{\sin}{x} + \epsilon$, $y_2 = \Fun{\sin}{x} - 2 \Fun{\exp}{-\sfrac{1}{2} \cdot (x-2)^2} + \epsilon$ and $y_3 = -1 - \sfrac{3}{8\pi} \cdot x + \sfrac{3}{10} \cdot \Fun*{\sin}{2x} + \epsilon$ with additive independent noise $\epsilon \sim \Gaussian*{0, 0.005^2}$.
The resulting data set $\D = \Set{\left( x, y_1 \right), \left( x, y_2 \right), \left( x, y_3 \right)}$ is trimodal in the interval $[0, 5]$ and is otherwise bimodal with one mode containing double the amount of data than the other.

We use squared exponential kernels as priors for both the $f^{\pix{k}}$ and $\alpha^{\pix{k}}$ and $25$ inducing points in every GP.
\Cref{fig:semi_bimodal} shows the posterior of a DAGP with $K = 4$ modes applied to the data, which correctly identified the underlying functions.
The figure shows the posterior belief about the assignments $\mat{A}$ and illustrates that DAGP recovered that it needs only three of the four available modes to explain the data.
One of the modes is only assigned points in the interval $[0, 5]$ where the data is actually trimodal.

This separation is explicitly represented in the model via the assignment processes $\mat{\alpha}$ (bottom panel in \cref{fig:semi_bimodal}).
Importantly, DAGP does not only cluster the data with respect to the generating processes but also infers a factorization of the input space with respect to the relative importance of the different processes.
The model has disabled the mode $k = 2$ in the complete input space and has learned that the mode $k = 1$ is only relevant in the interval $[0, 5]$ where the three enabled modes each explain about a third of the data.
Outside this interval, the model has learned that one of the modes has about twice the assignment probability than the other one, thus correctly reconstructing the true generative process.
The DAGP is implicitly incentivized to explain the data using as few modes as possible through the likelihood term of the inferred $\mat{a_n}$ in \cref{eq:variational_bound}.
At $x = -10$ the inferred modes and assignment processes start reverting to their respective priors away from the data.

\subsection{Mixed Cart-pole Systems}
\label{subsec:cartpole}
\begin{table}[t]
    \centering
    \caption{
        \label{tab:cartpole}
        Results on the cart-pole data set.
        We report mean log likelihoods with their standard error for ten runs.
        The upper results are obtained by training the model on the mixed data set and evaluating it jointly (left) on multi-modal predictions.
        We evaluate the two inferred sub-models for the default system (center) and short-pole system (right).
        We provide gray baseline comparisons with BNN+LV and GPR models which cannot solve the data assignment problem.
        BNN+LV yields joint predictions which cannot be separated into sub-models.
        Specialized GPR models trained the individual training sets give a measure of the possible performance if the data assignment problem would be solved perfectly.
    }%
    \sisetup{
        table-format=-1.3(3),
        table-number-alignment=center,
        separate-uncertainty=true,
        table-figures-uncertainty=1,
        detect-weight,
    }
    \newcolumntype{H}{>{\setbox0=\hbox\bgroup}c<{\egroup}@{}}
    \setlength{\tabcolsep}{1pt}
    \begin{tabular}{HlSSSS}
        \toprule
      & & \multicolumn{2}{c}{Mixed} & {Default only} & {Short-pole only} \\
      \cmidrule(lr){3-4} \cmidrule(lr){5-5} \cmidrule(lr){6-6}
        Runs & & {Train} & {Test} & {Test} & {Test} \\
        \midrule
        10 & DAGP & \bfseries 0.575 \pm 0.013 & \bfseries 0.521 \pm 0.009 & 0.844 \pm 0.002 & \bfseries 0.602 \pm 0.005 \\
        10 & DAGP 2 & 0.548 \pm 0.012 & \bfseries 0.519 \pm 0.008 & \bfseries 0.859 \pm 0.001 & 0.599 \pm 0.011 \\
        10 & DAGP 3 & 0.527 \pm 0.004 & 0.491 \pm 0.003 & 0.852 \pm 0.002 & 0.545 \pm 0.012 \\
        \addlinespace
        10 & OMGP & -1.04 \pm 0.02 & -1.11 \pm 0.03 & 0.66 \pm 0.02 & -0.81 \pm 0.12 \\
        \midrule
        \rowcolor{sStone!33}
        10 & BNN+LV & 0.519 \pm 0.005 & 0.524 \pm 0.005 & {\textemdash} & {\textemdash} \\
        \rowcolor{sStone!33}
        10 & GPR Mixed & 0.452 \pm 0.003 & 0.421 \pm 0.003 & {\textemdash} & {\textemdash} \\
        \rowcolor{sStone!33}
        10 & GPR Default & {\textemdash} & {\textemdash} & 0.867 \pm 0.001 & -7.54 \pm 0.14 \\
        \rowcolor{sStone!33}
        10 & GPR Short & {\textemdash} & {\textemdash} & -5.14 \pm 0.04 & 0.792 \pm 0.003 \\
        \bottomrule
    \end{tabular}
\end{table}
Our third experiment is based on the cart-pole benchmark for reinforcement learning as described by~\textcite{barto_neuronlike_1983} and implemented in OpenAI Gym~\parencite{brockman_openai_2016}.
In this benchmark, the objective is to apply forces to a cart moving on a frictionless track to keep a pole, which is attached to the cart via a joint, in an upright position.
We consider the regression problem of predicting the change of the pole's angle given the current state of the cart and the action applied.
The current state of the cart consists of the cart's position and velocity and the pole's angular position and velocity.
To simulate a dynamical system with changing system characteristics our experimental setup is to sample trajectories from two different cart-pole systems and merging the resulting data into one training set.
The task is not only to learn a model which explains this data well, but to solve the association problem introduced by the different system configurations.
This task is important in reinforcement learning settings where we study systems with multiple operational regimes.

We sample trajectories from the system by initializing the pole in an almost upright position and then applying 10 uniform random actions.
We add Gaussian noise $\epsilon \sim \Gaussian*{0, 0.01^2}$ to the observed angle changes.
To increase the non-linearity of the dynamics, we apply the action for five consecutive time steps and allow the pole to swing freely instead of ending the trajectory after reaching a specific angle.
The data set consists of 500 points sampled from the \emph{default} cart-pole system and another 500 points sampled from a \emph{short-pole} cart-pole system in which we halve the mass of the pole to 0.05 and shorten the pole to 0.1, a tenth of its default length.
This short-pole system is more unstable and the pole reaches higher speeds.
Predictions in this system therefore have to take the multimodality into account, as mean predictions between the more stable and the more unstable system can never be observed.
We consider three test sets, one sampled from the default system, one sampled from the short-pole system, and a mixture of the two.
They are generated by sampling trajectories with an aggregated size of 5000 points from each system for the first two sets and their concatenation for the mixed set.

For this data set, we use squared exponential kernels for both the $f^{\pix{k}}$ and $\alpha^{\pix{k}}$ and 100 inducing points in every GP.
We evaluate the performance of deep GPs with up to three layers and squared exponential kernels as models for the different functions.
As described in~\parencite{salimbeni_doubly_2017,kaiser_bayesian_2018}, we use identity mean functions for all but the last layers and initialize the variational distributions with low covariances.
We compare our models with OMGP and three-layer relu-activated Bayesian neural networks with added latent variables (BNN+LV).
The latent variables can be used to effectively model multimodalities and stochasticity in dynamical systems for model-based reinforcement learning~\parencite{depeweg_decomposition_2018}.
We also compare DAGP to three kinds of sparse GPs (GPR)~\parencite{hensman_scalable_2015}.
They are trained on the mixed data set, the default system and the short-pole system respectively and serve as a baseline comparison as these models cannot handle multi-modal data.

\Cref{tab:cartpole} shows results for ten runs of these models.
The GPR model predicts a unimodal posterior for the mixed data set which covers both systems.
Its mean prediction is approximately the mean of the two regimes and is physically implausible.
The DAGP and BNN+LV models yield informative multi-modal predictions with comparable performance.
In our setup, OMGP could not successfully solve the data association problem and thus does not produce a useful joint posterior.
The OMGP's inference scheme is tailored to ordered one-dimensional problems.
It does not trivially translate to the 4D cart-pole problem.

As BNN+LV does not explicitly solve the data association problem, the model does not yield sub-models for the two different systems.
Similar results would be obtained with the MDN and RGPR models, which also cannot be separated into sub-models.
OMGP and DAGP yield such sub-models which can independently be used for predictions in the default or short-pole systems.
Samples drawn from these models can be used to generate physically plausible trajectories in the respective system.
OMGP fails to model the short-pole system but does yield a viable model for the default system which evolves more slowly due to higher torque and is therefore easier to learn.
In contrast, the two sub-models inferred by DAGP perform well on their respective systems, showing that DAGP reliably solves the data association problem and successfully avoids model pollution by separating the two systems well.
Given this separation, shallow and deep models for the two modes show comparable performance.
The more expressive deep GPs model the default system slightly better while sacrificing performance on the more difficult short-pole system.

\section{Conclusion}
\label{sec:conclusion}
We have presented a fully Bayesian model for the data association problem.
Our model factorises the observed data into a set of independent processes and provides a model over both the processes and their association to the observed data.
The data association problem is inherently ill-constrained and requires significant assumptions to recover a solution.
In this paper, we make use of interpretable GP priors allowing global a priori information to be included into the model.
Importantly, our model is able to exploit information both about the underlying functions and the association structure.
We have derived a principled approximation to the marginal likelihood which allows us to perform inference for flexible hierarchical processes.
In future work, we would like to incorporate the proposed model in a reinforcement learning scenario where we study a dynamical system with different operational regimes.

\printbibliography

\end{document}